# Foot Anthropometry Device and Single Object Image Thresholding

Amir Mohammad Esmaieeli Sikaroudi[1], Sasan Ghaffari[2], Ali Yousefi[3], Hassan Sadeghi Naeini[4]

[1]Department of Industrial Engineering,
Iran University of Science and Technology, Tehran, Iran
[2,3,4]Department of Industrial Design,
Iran University of Science and Technology, Tehran, Iran

*ABSTRACT*

*This paper introduces a device, algorithm and graphical user interface to obtain anthropometric measurements of foot. Presented device facilitates obtaining scale of image and image processing by taking one image from side foot and underfoot simultaneously. Introduced image processing algorithm minimizes a noise criterion, which is suitable for object detection in single object images and outperforms famous image thresholding methods when lighting condition is poor. Performance of image-based method is compared to manual method. Image-based measurements of underfoot in average was 4mm less than actual measures. Mean absolute error of underfoot length was 1.6mm, however length obtained from side foot had 4.4mm mean absolute error. Furthermore, based on t-test and f-test results, no significant difference between manual and image-based anthropometry observed. In order to maintain anthropometry process performance in different situations user interface designed for handling changes in light conditions and altering speed of the algorithm.*

## 1. INTRODUCTION

Anthropometry plays a significant role in designing of the wearable products. Manufacturers of apparel products require a vast amount of anthropometric data to introduce a product which fits most customers. The scales and dimensions of a shoe are critical for customer convenience. However, a tradeoff between beauty and ergonomic factors is required. Considering customers with various genetic properties, the importance of anthropometric data emerged. Generally anthropometries are done in three main ways: manual anthropometry, 2D photography, and 3D scan. Based on usage, each method may have sufficient accuracy followed by its restrictions. Traditional methods of anthropometry has considerable difficulties and deficiencies.

In manual anthropometry or direct anthropometry a well-trained operator records the measurements using caliper and tape tools. This method is prone to poor accuracy caused by operator and tool errors. The error sources consist of, locating landmarks, the pressure applied, mechanical errors, posture and movement, and inter-observer errors [1-3]. Recently most of the anthropometries are implemented by image processing or software based methods.

Hung, Witana [4] considered the deficiencies of online apparel merchandising. Using a Visual Basic program coded specifically for the process, a digital camera and a 15 cm * 15 cm calibration square to scale the images, six linear and four circumferential measurements were obtained. Image based anthropometry was implemented by manual identification of landmarks on images by two operators. Image based method and manual method in three of ten measurements had significant difference. Error sources of image based method reported as difficulty in identification of body landmarks, shape quantifications and calibration which training of operators and knowledge of different body shapes can reduce error. Performance of image-based anthropometry in cloth sizing system is measured by Meunier and Yin [5]. In this article, clothing sizing has been aimed to work on as an issue which requires precise anthropometric data. Using a database of 349 subjects measured by both traditional and 2D image-based methods, and by comparing the results taken from each method, the authors concluded that image-base systems were capable of providing anthropometric measurements which were quite comparable to traditional measurement methods. Calibration of images was implemented by calibration markers and crosshairs on body were identified automatically as landmarks. Sources of errors in image based anthropometry errors were listed as perspective distortion, camera resolution, camera calibration, landmarking error and modelling error. BenAbdelkader and Yacoob [6] presented a geometric method to approximately calculate 3D line segments in an uncalibrated image. Landmarks on image were manually selected and some assumptions were used such as sufficient distance of person to assume planner distances. Anthropometric measurements were calculated in comparison to each other instead of absolute values hence image didn't require calibration. Final anthropometric values were estimated by statistical properties of lengths ratios. Gittoes, Bezodis [7] considered image based method as less time-consuming approach of measurements for obtaining anthropometric measurements for inertia modeling of athletic performers by implementing inertia model of Yeadon [8]. Six landmarks were set in image for calibration and presented method was implemented on five athletic male performers. Image-based method achieved similar accuracy in comparison to direct measurements. Li, Jia [9] presented a system which reconstructs 3D surface by multi-view images of an object rotating on a disk. This method requires a calibrated camera and focused on obesity and fitness measurements. 3D surface was constructed by graph cut algorithm and image consistency methods.

Xiong, Goonetilleke [10] obtained foot model for designing of good-fitted footwear on fifty Hong Kong Chinese adults. Simple measurements like length and width of foot were measured manually, but midfoot heights were measured by 3D scanner and landmarking. Relationship of length, height and ball girth were evaluated, however no direct relationship were discovered. Witana, Xiong [11] manually measured foot dimensions and compared it to 3D scan of foot. Foot scans were corrected in alignment for better performance and extracted measurements such as foot length, width, height and girth. It was reported that manual and 3D scan methods in 17 of 18 foot dimensions had no significant differences. Agić, Nikolić [12] foot measurements were extracted by 3d scan and relations of foot length, foot width, joint girth, hill circumference and stature were extracted by 103 Croatian adults. Furthermore, body mass index and arch index were implemented on underfoot. Foot shape differences focusing on sex and fringe size researched by 3D scanning and virtual landmarking of scanned models [13]. Distinction between male and female foot dimensions on Taiwanese adults researched by 3D scanning of placed landmarks on subject's foot. Dimensions classified into three groups based on gender and ethnic utilizing t-test and principal component analysis [14].

## 1.1 Image Thresholding

Image thresholding is an image processing technique to convert grey-scale image to binary image. Image thresholding is categorized to histogram shape-based methods, clustering-based methods, entropy-based methods, object attribute-based methods, the spatial methods and local methods [15]. Presented algorithm aims on a measurement called noise. Noise is defined in binary image that each binary image is obtained by a threshold. All possible thresholds are tested with respect to two constraints. The threshold that has minimum noise is selected as optimum threshold and a post-process of edge noise removal is implemented. This method is compared to Huang's fuzzy thresholding method [16], Intermodes or middle point of bimodal histogram peaks [17], IsoData [18], Li's Minimum Cross Entropy thresholding method [19], Maximum Entropy thresholding method [20], mean of grey levels as the threshold [21], iterative Minimum Error thresholding [22], Minimum or minimum point between two peaks of bimodal histogram [17], Tsai's moments method [23], Otsu's threshold clustering algorithm [24], Percentile [25], RenyiEntropy [25], Shanbhag [26], Triangle method [27] and Yen's thresholding method. Results shows that mentioned methods have good performance in easy cases, but as lighting situation deteriorates none of them was stable in results.

## 2. METHOD

This experimental case study has two phases in which in the first step a new practical device was developed to introduced economic equipment instead of a high tech 3D scanning. Furthermore, an image processing algorithm is developed which was embedded in a user interface. The aim in this research is reliability, hence obtaining optimum results is preferred to real-time processing. In the 2nd phase 15 males 20-25 years old samples were measured by the mentioned method to clarify its efficiency. In order to compare manual and image-based methods, statistical tests were performed.

## 2.1 Device

This research has a physical apparatus containing an opaque glass in floor, a white background with LED lights which are located between legs, a mirror under the opaque glass and a digital camera in front of them. Additionally a shutter is used to avoid touching camera and two sources of light might be required for front and back of the foot if environment light is low or light of front and back of foot are not equal. Two sources of light are equipped on white background if required and their height can be changed to configure intensity of light. Mirror provides image of underfoot simultaneously. This apparatus is illustrated in figure 1.

After retrieving foot images, they are fed into image processing algorithm by a simple graphic user interface (GUI). This GUI has tuning parameters which affect calculations amount/accuracy and adapts the algorithm with environment lighting situation. Furthermore, GUI is designed for mass processing and saving of results if the digital camera memory is full. Measurements of foot are calculated by number of pixels and pixels are converted to real length by scaling function.

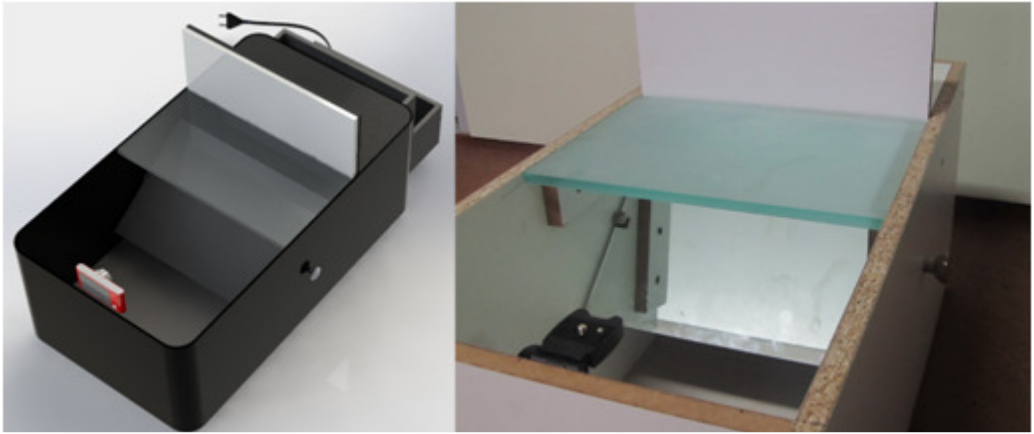

Fig. 1. Apparatus for gathering data to feed image processing algorithm.

Scaling function specifies number of pixels which one centimeter contains according to distance of foot from white background or camera. Distance of foot from white background is specified by underfoot image. In order to obtain scaling function a blue cube with five centimeters length in each dimension was used. This cube is set on different distances and based its number of pixels a linear function was fitted. This process was implemented once for each apparatus. Scaling function is illustrated in figure 2 and the blue cube showed in figure 3. For instance, if an object has 300 pixels distance to background, which is obtained from under image, then each centimeter will contain about 120 pixels for side foot and about 80 pixels for underfoot.

Although fixed camera was an assumption is this research, but LEDs provide a bright section which facilitates underfoot and side foot separation. However, a red label on opaque glass can provide this separation. Figure 4 illustrates an image taken before performing any process.

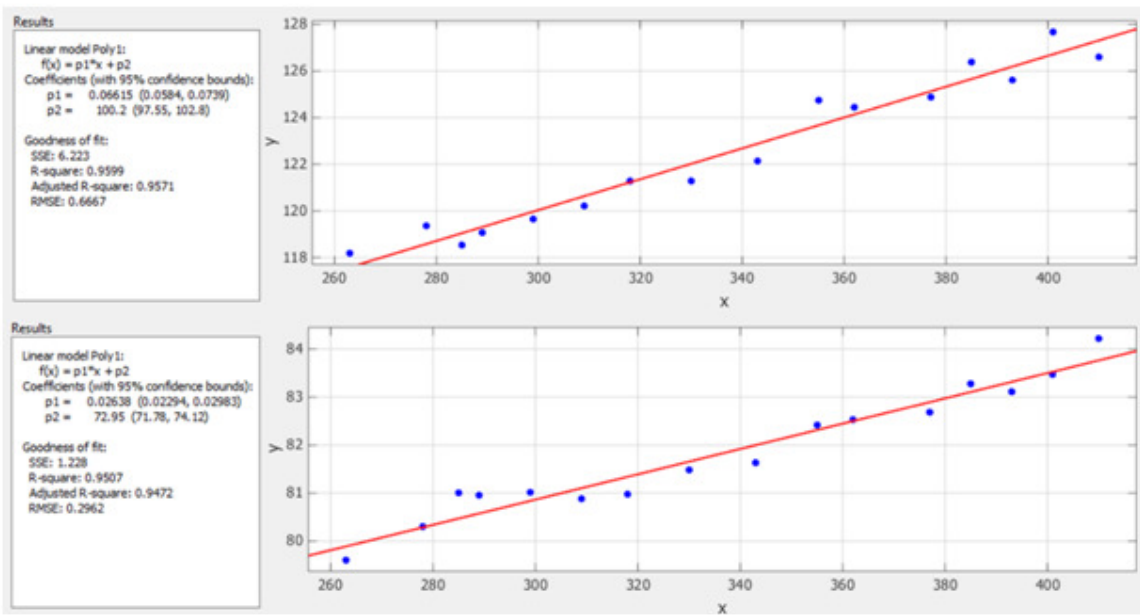

Fig. 2. Upper function is scaling function of side foot image and lower is for underfoot image

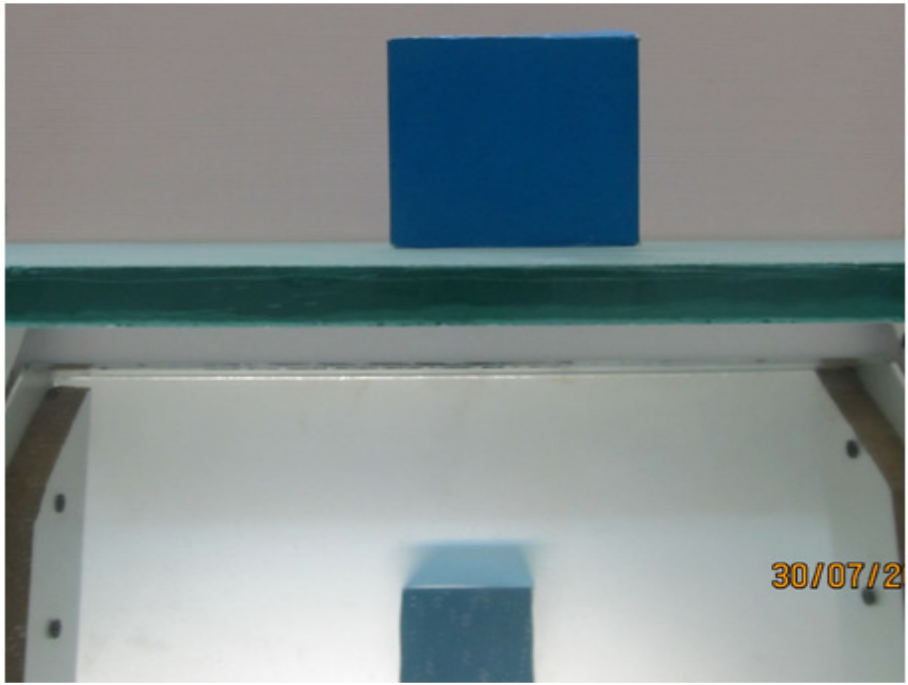

Fig. 3. A cube used with 5 centimetres in each dimension used to make scale function.

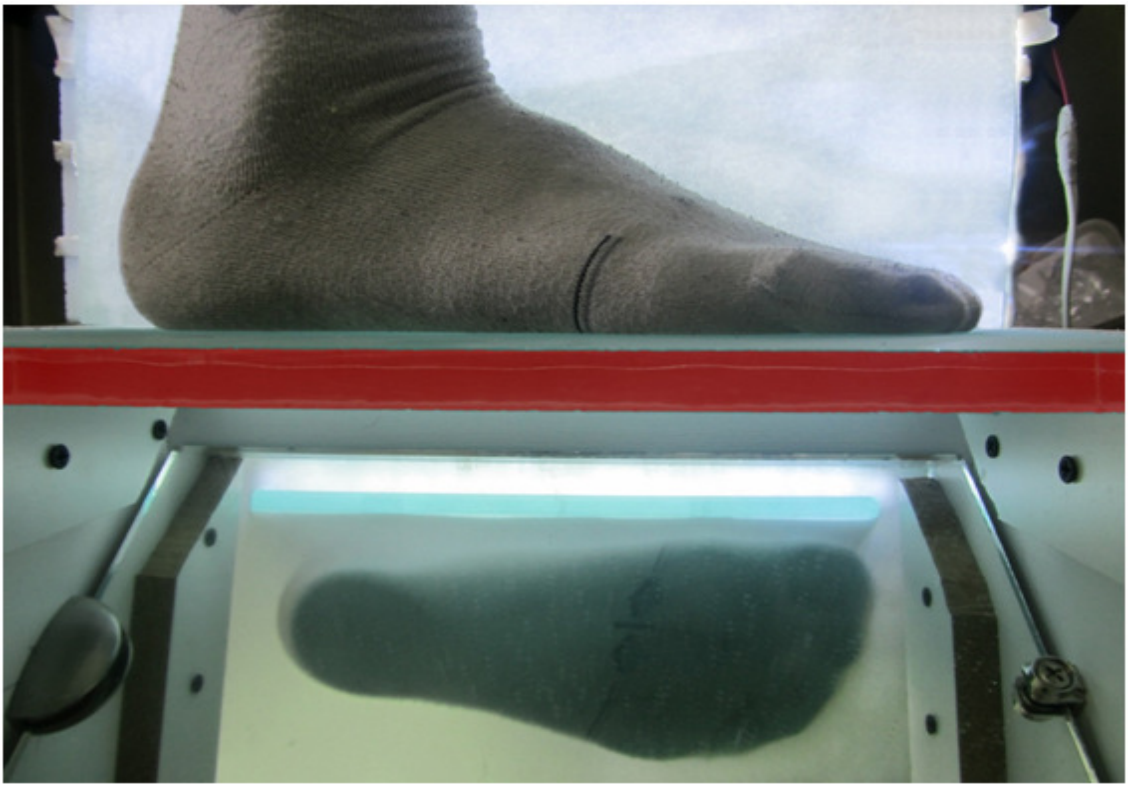

Fig. 4. Raw image taken by camera.

## 2.2. Side foot processing

Background light can be estimated manually or by averaging of sampled pixels of background. To specify foot length from side foot image, from the start of white background, column by column and from top to down, brightness of the pixels are compared to the background brightness. If pixel's value is significantly less than background's value, the pixel is considered as the foot starting pixel. After finding the starting pixel, if all pixels of a column are as bright as the background, this column of pixels is considered as the end of foot.

In order to calculate foot height based on the half of foot length, in the half foot length column of pixels, from bottom of image anywhere had minimal brightness is selected as bottom of foot and as reached to upper bright pixels of the background it is selected as peak of the foot. By subtraction bottom of foot pixel from upper foot pixel in half of foot column of pixels and converting by scaling function, foot height will be specified. Environment light conditions may not form a suitable shadow under the foot, hence in this situation two sources of light should be mounted on background.

Further information can be obtained from foot image such as upper foot curve. In order to find initial point of curve, gradients of pixels are required. Gradient of pixels is defined as the height difference of two pixels in adjacent column of pixels. In order to initiate the upper foot curve, if gradient of pixels is zero, it means that the algorithm is in ankle part of the image (i.e. top edge of the image), if positive, the algorithm is in back curve of the foot and if negative, the algorithm is in fore cure of the foot. As the starting point of foot fore curve is specified, the algorithm moves column by column and from top to down. If a pixel was different from the background it's saved as one point of upper curve. This process continues for a portion of foot length and saves position of pixels which is illustrated in figure 5.

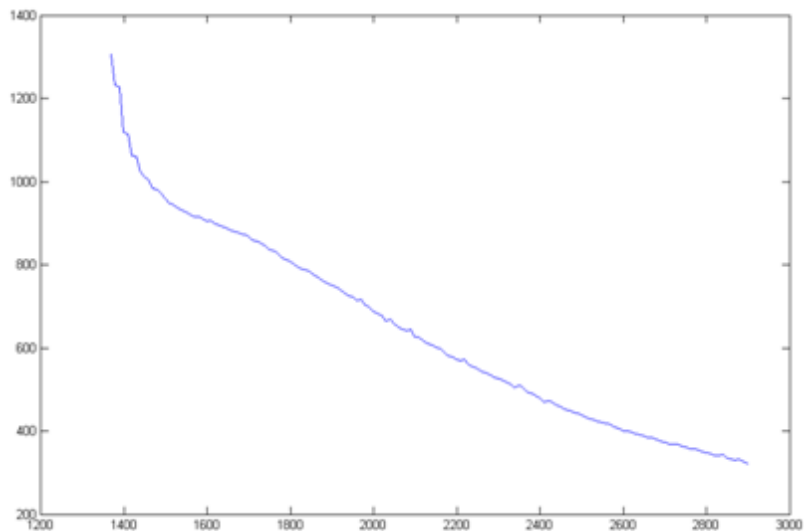

Fig. 5. Foot upper curve extracted from image.

## 2.3. Underfoot processing

Processing underfoot image requires a search for best threshold in gray-scaled image to accept pixels as underfoot. Each threshold results a binary image that defines accepted and not accepted pixels. In order to find the optimum threshold, in every step of threshold search, noise should be calculated and the threshold with minimum noise is accepted. Noise in binary image is divided into the noise in rows of pixels and the noise in columns of pixels. Noise in a row is the number of changes from the not accepted pixels to the accepted pixels and vice versa. Total noise is the average of the noises in rows and the noises in columns. This process is suitable for images which only contain one target object, therefore it's called single object image thresholding (SOIT).

Threshold search in first and final values will result zero noise because no pixels are selected or all pixels are selected. The main constraint for avoiding this issue is that an underfoot image has limits on minimum and maximum number of pixels in the taken image. This constraint can be from 20 percent to 70 percent of the image pixels. In addition, any accepted pixel in edge of the image should be considered as noise. Figure 6 illustrates entire threshold search which the red area is infeasible according to the constraint of number of pixels. Finally, the threshold number with minimum noise can determine underfoot object, but it may contain noise in underfoot edge or anywhere in image. Noise can be reduced as post processing section by removing accepted pixels with low adjacent accepted pixels and accepting none-accepted pixels with high adjacent accepted pixels. Each pixel has eight adjacent pixels. Figure 7 is the result of optimum threshold which noise reduction is implemented on it. SOIT formulation is expressed as below:

$$Minimize\ z = \left(\left(\sum_{i=1}^{IW} NR_i \Big/ IW\right) + \left(\sum_{j=1}^{IH} NC_j \Big/ IW\right)\right)\Big/2 + 20*(EN/(IW+IH))$$

Subjected to:

$$\frac{NAC}{IW*IH} > 0.2$$

$$\frac{NAC}{IW*IH} < 0.7$$

Where:

$NR_i = Noise\ in\ row\ i$

$NC_j = Noise\ in\ column\ j$

$EN = Edge\ noise$

$IW = Image\ width$

$IH = Image\ height$

$NAP = Number\ of\ accepted$ pixels

If SOIT is used for detection of multiple objects, it tries to remove smaller objects to reduce noise criteria. However, if constraints on number of pixels are adjusted properly and size of objects be similar to each other, this algorithm can perform well on multiple object detection.

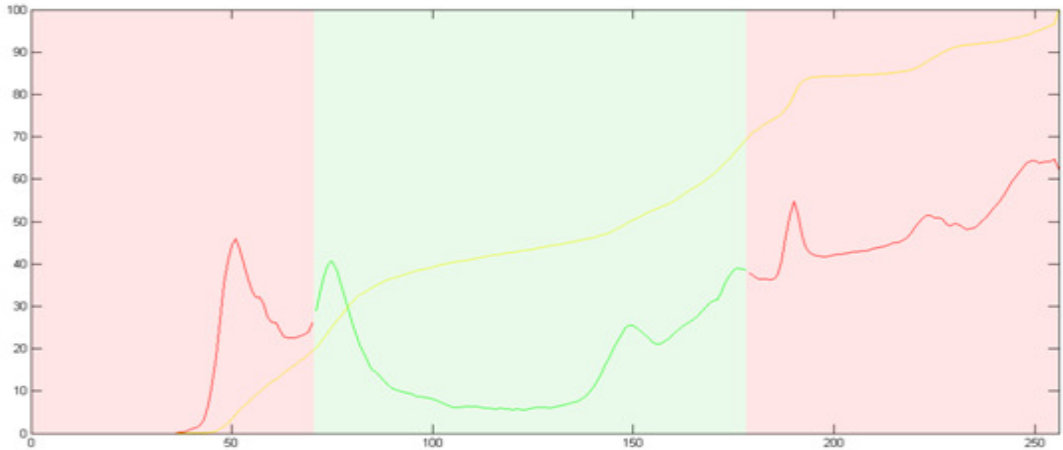

Fig. 6. Noise in threshold search with respect to percent of accepted pixels constraint. Red lines: infeasible area of objective function, green line: feasible area of objective function, yellow line: percent of accepted pixels

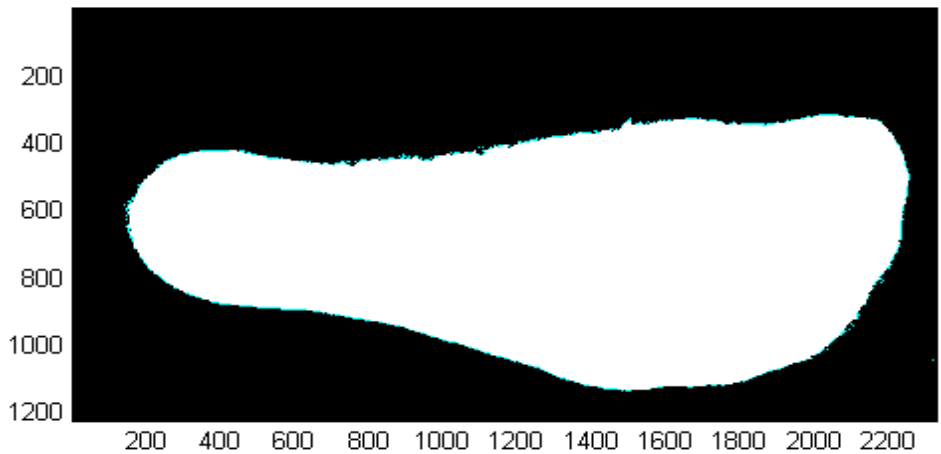

Fig. 7. Final underfoot image, blue pixels are discarded to reduce noise.

### 2.4. Software

This research is codded in Java SE. In order to track algorithm process, GUI illustrated in figure 8 is developed. Figures were developed in Java in order to monitor error plot of algorithm. If user wants to ignore automatic background brightness sampling, background brightness could be determined manually. User can select suitable background brightness according to location of the device. Furthermore, user defines bounds of threshold search and search steps to increase algorithm speed. If user ensured about settings based on device location, algorithm could be set for mass calculation without refreshing plots to increase outputting speed.

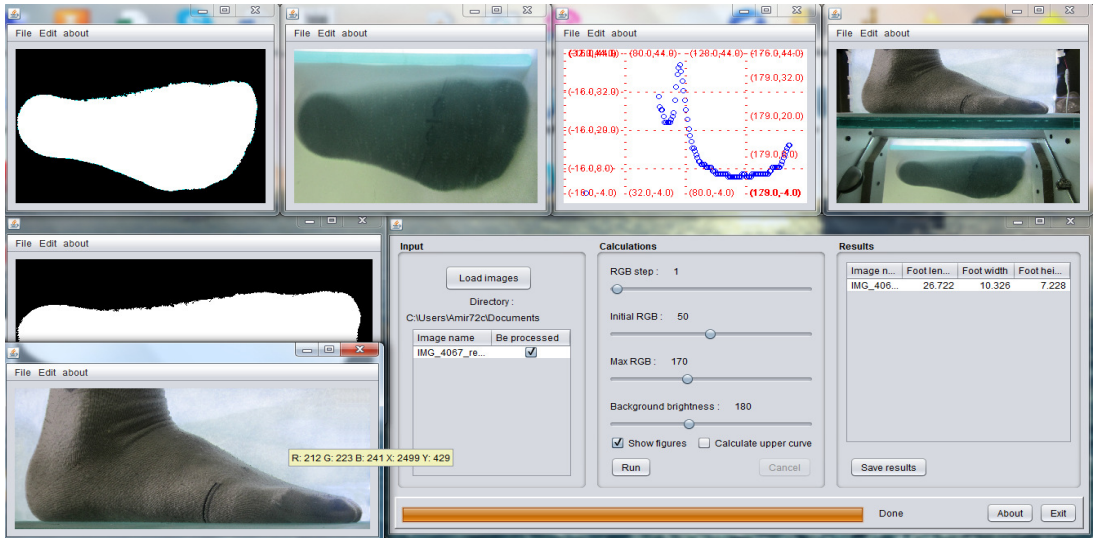

Fig. 8. Graphic user interface designed to set tuning parameters and mass calculation of images.

## 3. RESULTS

In order to validate presented image-based method, foot measurements of 15 males 20-25 years old were obtained by manual and image-based method. Results are shown in table 1.

Table 1. Measurement results of manual and image-based methods.

| | Manual measurements (cm) | | | Image-based measurements (cm) | | | |
|---|---|---|---|---|---|---|---|
| Number | Length | Width | Height | Length side | Length under | Height | Width |
| 1 | 27.5 | 10 | 7.9 | 27.20293 | 27.08039 | 8.381489 | 9.750772 |
| 2 | 26.3 | 10.5 | 7.2 | 26.41688 | 26.18321 | 7.688411 | 10.31316 |
| 3 | 26.6 | 10.5 | 7 | 25.90345 | 26.32011 | 7.131658 | 10.80731 |
| 4 | 27.5 | 10.6 | 6.7 | 28.03944 | 27.65997 | 6.934856 | 10.40141 |
| 5 | 26.7 | 10.3 | 6.3 | 25.99526 | 26.59397 | 6.010266 | 10.27536 |
| 6 | 27.5 | 10.6 | 6.4 | 27.11921 | 27.22144 | 6.765132 | 10.46038 |
| 7 | 28 | 10.4 | 7.4 | 28.34077 | 28.10392 | 7.429675 | 10.58929 |
| 8 | 26.4 | 10.4 | 6.8 | 27.4176 | 26.55279 | 7.029969 | 10.71475 |
| 9 | 26.6 | 10.5 | 7 | 26.49505 | 26.7236 | 7.057933 | 9.919353 |
| 10 | 25.7 | 9.6 | 7.2 | 25.96723 | 25.96951 | 7.036451 | 9.472597 |
| 11 | 26.3 | 10.5 | 7.2 | 25.91764 | 26.23438 | 7.153327 | 10.44034 |
| 12 | 28 | 10.4 | 7.4 | 28.6438 | 28.10606 | 7.239411 | 10.62542 |
| 13 | 26.4 | 10.4 | 6.8 | 26.52991 | 26.62558 | 7.019659 | 10.6374 |
| 14 | 26.6 | 10.5 | 7 | 25.79124 | 26.72754 | 6.990498 | 10.49068 |
| 15 | 25.7 | 9.6 | 7.2 | 25.45895 | 25.69753 | 6.970006 | 9.901782 |
| | Mean absolute error | | | 0.444793 | 0.169197 | 0.209255 | 0.210125 |

Signal & Image Processing : An International Journal (SIPIJ) Vol.8, No.3, June 2017Signal & Image Processing : An International Journal (SIPIJ) Vol.8, No.3, June 2017

In device context, background of image should have uniform light. Therefore, a translucent cover on LEDs boosts algorithm performance. Camera must be fixed and accessible remotely. In order to validate process, t-test and f-test is implemented. P-value for significance of difference in results was calculated. Correlation of manual and image based measurement results was calculated by Pearson method. Table 2 summarized descriptive statistics and statistical tests. Image-based method had no measurement with significant difference to manual measurement. Although tests implied poor performance of side foot length, but height had high accuracy.

Table 2. Descriptive statistics and statistical analysis of results.

| Variable | Length side | Length under | Height | Width |
| --- | --- | --- | --- | --- |
| Mean | 26.749 | 26.787 | 7.123 | 10.320 |
| SE Mean | 0.258 | 0.187 | 0.129 | 0.100 |
| StDev | 0.999 | 0.726 | 0.499 | 0.388 |
| Minimum | 25.459 | 25.698 | 6.010 | 9.473 |
| Q1 | 25.918 | 26.234 | 6.970 | 9.919 |
| Median | 26.495 | 26.626 | 7.036 | 10.440 |
| Q3 | 27.418 | 27.221 | 7.239 | 10.625 |
| Maximum | 28.644 | 28.106 | 8.381 | 10.807 |
| t-test, p-value | 0.454 | 0.500 | 0.500 | 0.297 |
| f-test, p-value | 0.139 | 0.467 | 0.260 | 0.222 |
| Pearson correlation | 0.851 | 0.961 | 0.871 | 0.750 |

Raw data obtained by image-based method from underfoot image both in length and width are 4mm less than manual method in average. Penumbra around the foot and pixels removal of foot edge result in less than actual values. Background LEDs in addition to providing brightness of side foot background, they provide a gap between foot and background in underfoot image. If user set the foot close to background, shadow in the gap is minimized by background LEDs. Therefore, algorithm can separate the foot from narrow gap in the upper edge of the underfoot image.

Results of SOIT are compared to other methods of image thresholding provided in Fiji Auto Threshold plugin used in ImageJ software. Figure 9, 10 and 11 compare SOIT and other methods in different lighting conditions.

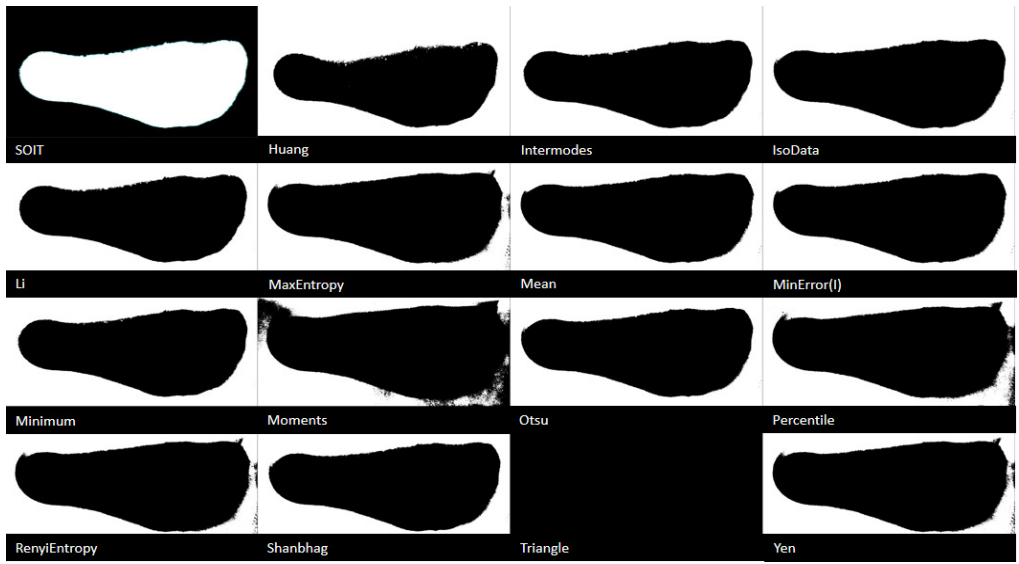

Fig. 9. Comparison of models in good lighting conditions.

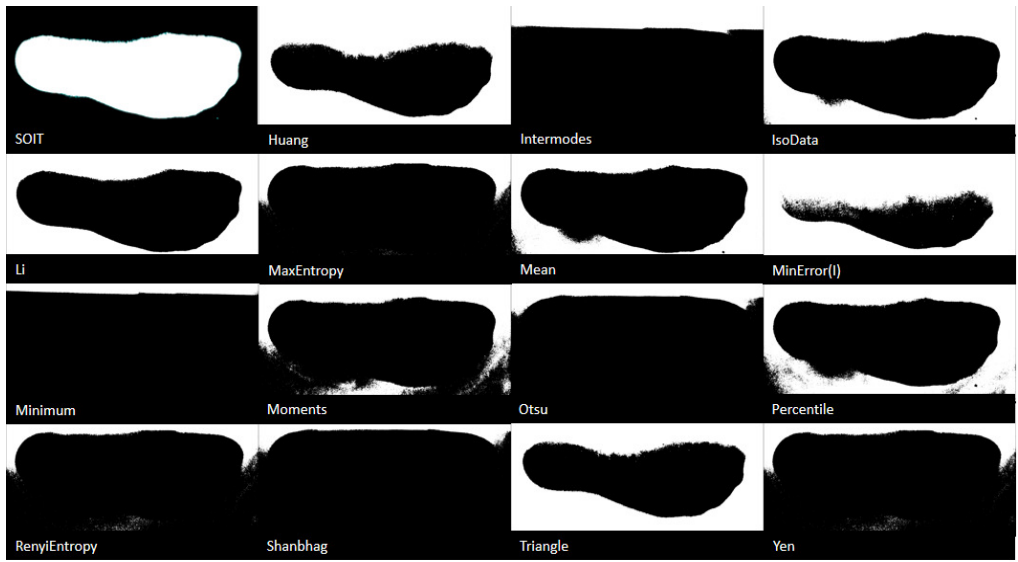

Fig. 10. Comparison of models in average lighting conditions.

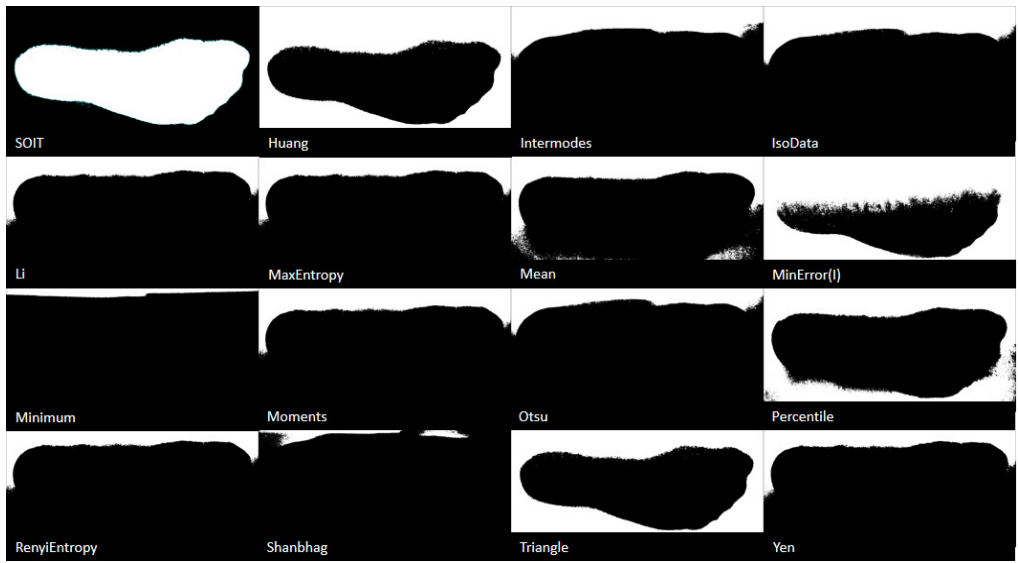

Fig. 11. Comparison of models in poor lighting conditions.

## 4. CONCLUSION

This research introduced a low cost anthropometry device and an algorithm which outperforms famous image thresholding methods. SOIT algorithm is more accurate and reliable in various conditions, however it's slower than other methods and it's not a general purpose algorithm because it can perform well on single object images. In order to speed up the algorithm, searching area can be reduced by user in GUI. Opacity of underfoot glass removes farther objects and boosts SOIT performance because searching of threshold in noisy image gradually adds/removes objects which guides SOIT algorithm to detect the object. Hence, an image which has no noise may not be suitable for SOIT algorithm. Presented process has low error compared to manual method, however results implies that underfoot measurements are more accurate. From poor performance of foot length from side foot image and good accuracy of foot height it can be inferred that the camera lens has impact on the results. Although simultaneous images of underfoot and side foot mitigates scaling errors, but considerations of the camera lens should be studied for achieving better accuracy.